\documentclass[pdflatex,sn-mathphys-num]{sn-jnl}

\usepackage{graphicx}%
\usepackage{multirow}%
\usepackage{amsmath,amssymb,amsfonts}%
\usepackage{xcolor}%
\usepackage{textcomp}%
\usepackage{booktabs}%
\usepackage{tikz}%
\usetikzlibrary{arrows.meta}%

\begin{document}

\title[HERON]{HERON: Real-Time Obstacle-Aware Reachability Maps for Mobile Manipulators}

\author[1]{\fnm{Guillaume} \sur{Dupoiron}}\email{guillaume.dupoiron.1@ens.etsmtl.ca}

\author[2]{\fnm{Fran\c{c}ois} \sur{Michaud}}\email{francois.michaud@usherbrooke.ca}

\author[1]{\fnm{Jean-Philippe} \sur{Roberge}}\email{jean-philippe.roberge@etsmtl.ca}

\affil[1]{\orgdiv{Command and Robotics Laboratory}, \orgname{\'Ecole de technologie sup\'erieure}, \orgaddress{\city{Montreal}, \postcode{H3C 1K3}, \state{Quebec}, \country{Canada}}}

\affil[2]{\orgdiv{Interdisciplinary Institute for Technological Innovation (3IT), Department of Electrical Engineering and Computer Engineering}, \orgname{Universit\'e de Sherbrooke}, \orgaddress{\city{Sherbrooke}, \postcode{J1K 0A5}, \state{Quebec}, \country{Canada}}}

\abstract{Mobile manipulator robots rarely reposition their base during manipulation, because doing so requires real-time knowledge of how well the arm can reach its surroundings among obstacles. Reachability maps (RMs) encode this knowledge, but existing methods are slow, coarse (typically 5 cm), or ignore real obstacles. This article asks whether a neural network can predict an obstacle-aware RM at 2 cm resolution from a live depth camera, on embedded hardware, fast enough for online use. We present HERON, a 3D U-Net trained on RMs computed with the GPU-accelerated inverse kinematics of \texttt{cuRobo}, first on synthetic scenes and then on real depth data, and deployed on a Jetson AGX Orin with TensorRT and INT8 quantization. HERON predicts the RM of a 2.56 m cube around the arm in 115 ms on average, for a measured system rate of 8.57 Hz, close to the 10 Hz update rate of the obstacle map, whereas computing the same map with \texttt{cuRobo} takes about 90 min. Against the \texttt{cuRobo} ground truth, 95\% of the voxels lie within $\pm$1 class out of 51 with the FP32 model and within $\pm$3 classes with the INT8 model, with a Pearson correlation of 0.99 and a quadratic weighted kappa above 0.9. Accuracy degrades mainly at the workspace boundaries.}


\maketitle

\section{Introduction}
\label{sec:Intro}

Mobile manipulator robots (MMs) combine the mobility of a mobile base with the dexterity of a robotic arm, offering a flexibility that fixed manipulators cannot match. To decide where to place its base, an MM needs to know how well its arm can reach the space around it. A reachability map (RM) provides this knowledge: it divides the space around the arm into voxels and scores each voxel by the fraction of end-effector orientations that the arm can reach there~\cite{Zacharias}. Humans rely on a similar, intuitive sense of what is reachable to position themselves before acting~\cite{rabellino_peripersonal_2020}.

Yet MMs are often implemented as an arm mounted on a base with little coordination between the two subsystems, which Ghodsian et al.~\cite{ghodsian_mobile_nodate} identify as one of the main barriers to their deployment in industry. Whole-body controllers coordinate base and arm motion while accounting for obstacles~\cite{Pankert2020}, and other approaches optimize physically motivated metrics such as manipulability or distance to joint limits~\cite{thakar_survey_2022}. However, none of them assesses the quality of the workspace in which they place the robot, with respect to the task and to the surrounding obstacles. For tasks with a large range of motion, such as cleaning or painting, the base must be repositioned as the task evolves, which requires this assessment online.

\begin{figure}[t]
    \centering
    \includegraphics[width=0.55\linewidth]{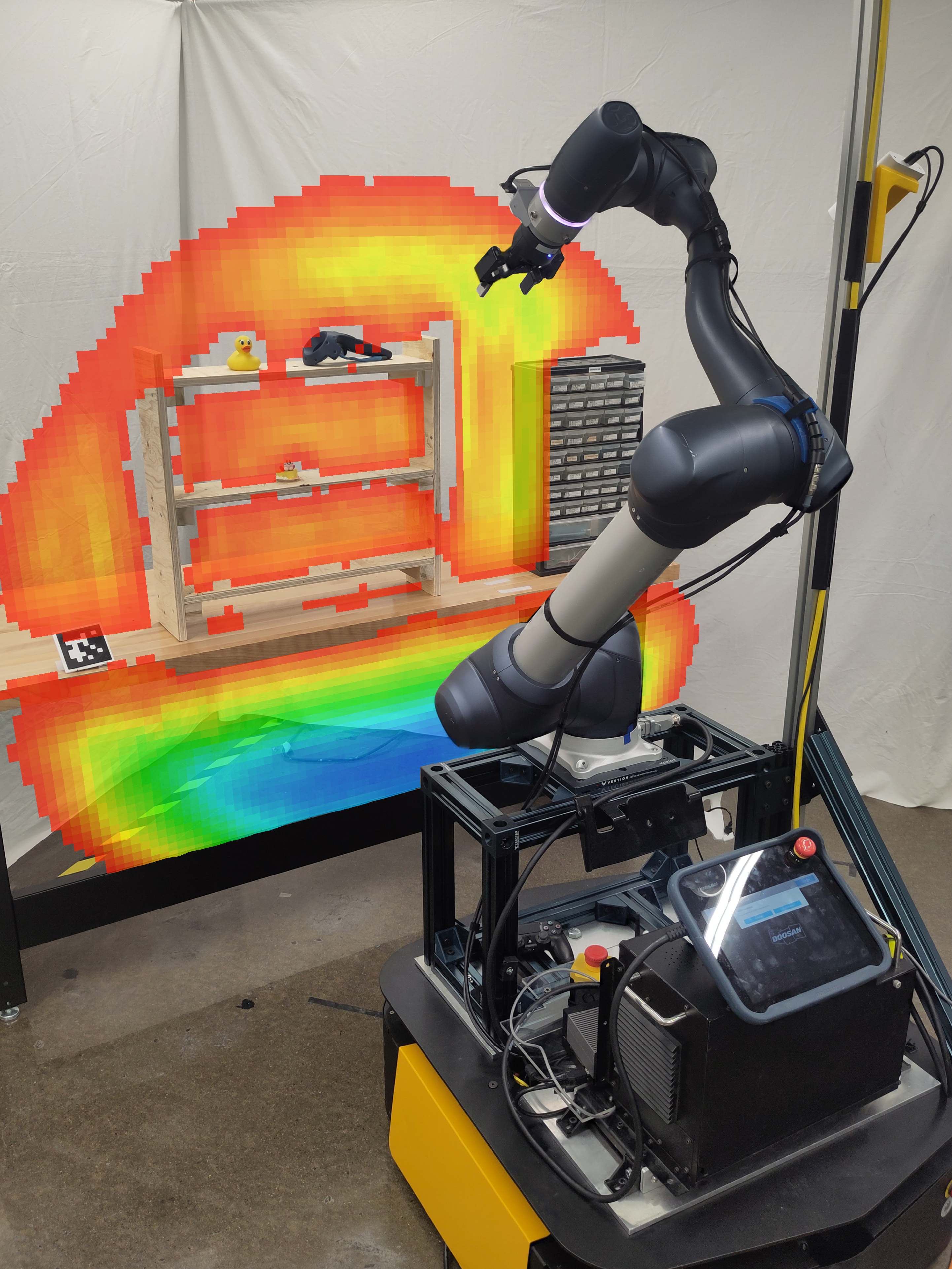}
    \caption{Experimental setup. HERON runs on a Jetson AGX Orin onboard an MM composed of a Clearpath Ridgeback omnidirectional base, a Doosan M1013 arm, and an Azure Kinect depth camera. A vertical slice of the generated RM is overlaid on the photograph, aligned with an ARTag marker.}
    \label{fig:setup_robot}
\end{figure}

Current RM methods cannot provide this assessment in real time: they are slow to compute, operate at coarse resolution, typically 5 cm, and largely ignore real obstacles. The cause is computational: evaluating inverse kinematics (IK) with collision checking for every pose of every voxel is prohibitively slow on CPU at fine resolution, and maps precomputed without obstacles become inaccurate near them.

Two recent GPU tools change this situation: \texttt{cuRobo}~\cite{Balakumar2023} solves batched IK with collision checking, and nvblox~\cite{millane_nvblox_2024} maintains a distance field of the scene from depth images. Together, they make it tractable to compute ground-truth RMs at 2 cm resolution in scenes with real obstacles. Computing one RM still takes about 90 min, far too slow for online use, but such maps can train a network that predicts the RM directly from an obstacle map.

This paper presents HERON, a High-resolution Embedded Reachability Obstacle-aware Network, and addresses the following question: can a neural network predict an obstacle-aware RM at 2 cm resolution from a live depth camera, on embedded hardware, fast enough for online base repositioning and close enough to the \texttt{cuRobo} ground truth? We answer it with two criteria: the prediction rate, which we aim to bring as close as possible to the 10 Hz update rate of the obstacle map so that the RM can later drive a closed-loop base controller, and the agreement with the \texttt{cuRobo} ground truth, measured in several regions of the workspace. We assume a single known arm, obstacles observed by a single depth camera, and a ground truth defined by \texttt{cuRobo} IK over 50 orientations per voxel. Each prediction reflects a single snapshot of the obstacle map, so obstacles that move during the approximately 120 ms of a prediction are not accounted for.

The paper is organized as follows. Section~\ref{sec:RelatedWork} reviews research on RMs and mobile base placement. Section~\ref{sec:Methodology} presents the dataset, the network, and the evaluation metrics. Section~\ref{sec:ExperimentalSetup} describes the hardware, the training, and the real-time implementation. Section~\ref{sec:Results} presents the results, Section~\ref{sec:Discussion} discusses them, and Section~\ref{sec:Conclusion} concludes.

\section{Related Work}\label{sec:RelatedWork}

Positioning an MM in a cluttered environment requires evaluating the robot's workspace, typically encoded as an RM. Over the past fifteen years, a substantial body of research has addressed this problem. Existing methods can be evaluated along four criteria: 1) the environment sampling strategy, 2) the metrics used to score the workspace, 3) the computational cost, and 4) the handling of obstacles. A fifth dimension, generalizability across robot morphologies,  has received comparatively less attention. 

Zacharias et al.~\cite{Zacharias} introduced one of the first voxel-based RM representations. Their initial approach sampled the joint space uniformly and associated configurations with voxels through forward kinematics (FK). However, as they noted: \textit{``When a robot is in a singular configuration, large steps in the configuration space for the links causing the singularity result in small motions in the Cartesian workspace''}, biasing the map toward singular configurations. The alternative based on IK inscribes a sphere in each voxel and evaluates the fraction of uniformly distributed poses that are reachable:
\begin{equation}
    R_i = \frac{R}{N}, \quad R \leq N
    \label{eq:reachability_index}
\end{equation}
where $N$ is the number of poses sampled in voxel $i$ and $R$ the number of them with a valid IK solution. Yao et al.~\cite{yao_enhanced_nodate} recently extended it to account for the redundancy of 7-DOF arms through an Enhanced Dexterity Map. To reduce the computational cost of IK-based generation, Porges et al.~\cite{porges_reachability_2013} proposed a hybrid approach that uses FK heuristics up to a threshold and fills remaining voxels with IK. Cavelli et al.~\cite{cavelli_modeling_2025} took a different route, approximating the reachability shape by fitting ellipsoids to joint-sampled point clouds, bypassing IK and FK entirely. While these methods improve generation speed, they all assume the robot base operates on a flat surface, an assumption challenged by Birr et al.~\cite{birr_oriented_2022}, who extended the RM framework to inclined support planes, highlighting that most approaches are misaligned with real unstructured environments.

The core limitation of RM-based methods is their computational cost, which prevents online use. Vahrenkamp et al.~\cite{Vahrenkamp2013} address this by inverting the problem: rather than asking ``which poses can I reach from this base position?'', their Inverse Reachability Map (IRM) asks ``where should the base be placed to reach this pose?'' By precomputing a database of valid base poses offline, the IRM enables fast online queries. While effective, this approach shifts rather than solves the original problem: the RM must still be precomputed in an obstacle-free configuration, and map accuracy degrades near obstacles at runtime. Makhal et al.~\cite{Makhal2018} improve IRM generation speed using IKFast~\cite{diankov_automated_2018} and introduced a union map for multi-target placement, providing an open-source ROS implementation that has since become a widely used baseline. Zhong et al.~\cite{zhong_optimization_nodate} fuse occupancy grid maps with the IRM to partially account for obstacles at query time, though accuracy remains limited near obstacles. More recently, Rudorfer~\cite{rudorfer_rm4d_2025} reduced the map to four dimensions for common 6- and 7-axis arms, combining the RM and the IRM in a single structure built an order of magnitude faster. This map nevertheless remains precomputed offline.

To achieve real-time performance, several approaches have reduced problem complexity at the cost of accuracy or applicability. Reister et al.~\cite{reister_combining_nodate} combine real-time RM queries with robot navigation for table-clearing tasks. Burgess-Limerick et al.~\cite{burgess-limerick_enabling_2023, burgess-limerick_reactive_2024} simplify the problem further by computing only 2D reachability maps, enabling fast online inference at the expense of 3D workspace detail. Both methods demonstrate real-time capability but only in simple, uncluttered environments where obstacle-induced accuracy degradation is minimal.

Learning-based methods have emerged as a promising direction to generate RMs or IRMs in milliseconds while handling obstacles. Chen et al.~\cite{chen_generating_nodate} propose combining a long short-term memory (LSTM) with a ResNet to generate capability maps for arbitrary robot shapes, but their model cannot handle self-collisions or environment collisions. Sandakalum et al.~\cite{sandakalum_inv-reach_nodate} update the IRM to account for obstacles with a two-stage approach: a classification model determines the IRM shape, then a regression model scores it. Through preprocessing, this approach is 250 times faster than Reuleaux~\cite{Makhal2018}. However, it is limited to IRM generation, operates at 5 cm resolution, and does not generalize across robot morphologies. Their subsequent work, ReachNet~\cite{sandakalum_reachnet_nodate}, extends the approach to direct, obstacle-aware RM generation with a regression network fast enough for real-time base placement, but only for a 3-DOF planar robot at 10 cm resolution, far below the granularity required in cluttered environments. Naik et al.~\cite{naik_basenet_2024} bypass map generation entirely using reinforcement learning to directly predict optimal base poses, but the resulting policy does not generalize across environments or robot configurations.

The shared bottleneck of all these methods is dataset generation: computing IK with collision checking at fine resolution is prohibitively slow on CPU, constraining most learning-based approaches to resolutions around 5 cm. Collision representation is a second bottleneck that limits map resolution in practice. Obstacles can be of two types: offline obstacles such as self-collisions, typically defined by the user as geometric primitives (cylinders, spheres, cubes) or meshes; and online obstacles reconstructed from depth cameras at runtime. Online obstacles are traditionally encoded as Octomap structures~\cite{hornung_octomap_2013} or occupancy grids, both of which rely on CPU-based computation and scale poorly to fine resolutions. At 2 cm resolution, the memory and computation costs of maintaining such representations in real time become prohibitive on standard hardware. Recently, Euclidean Signed Distance Fields (ESDF) have emerged as an alternative representation that encodes, for each voxel, the metric distance to the nearest obstacle surface. Millane et al.~\cite{millane_nvblox_2024} propose nvblox, a GPU-native ESDF implementation that incrementally updates the distance field from depth camera streams, shifting the collision computation bottleneck from CPU to GPU. This representation is directly compatible with gradient-based planners and enables collision checking at resolutions down to 1 cm with latencies compatible with real-time operation. Sundaralingam et al.~\cite{Balakumar2023} recently introduced \texttt{cuRobo}, a GPU-accelerated motion planning library that computes batched IK solutions in milliseconds and supports collision checking at 1 cm resolution. As a result, both dataset generation and online inference can now leverage GPU parallelism, making datasets at 2 cm resolution tractable.

Validating RMs and base placement methods remains a largely unsolved methodological challenge, as the contribution of map accuracy to task success is difficult to isolate. Validation practices in the literature fall into three levels of increasing rigor. The first is qualitative: visual inspection of the workspace and geometric consistency checks~\cite{Zacharias, porges_reachability_2015}. The second is comparative: performance against existing methods, measured by computation time or task success rate~\cite{Makhal2018, yao_enhanced_nodate}. The third is quantitative: errors such as the mean absolute error (MAE) or the root-mean-square error (RMSE) against an IK-generated reference~\cite{sandakalum_inv-reach_nodate, cavelli_modeling_2025}. A fourth level, which to the best of our knowledge no prior work has achieved, would require sampling the RM with a real robot in environments with obstacles and statistically representative sensor noise. It would provide the only ground truth that fully reflects deployment conditions.

Despite recent progress, to the best of our knowledge, no existing method simultaneously achieves real-time RM generation, 2 cm resolution, low data requirements, and generalizability across robot morphologies. 

\section{Methodology}
\label{sec:Methodology}

To address this gap, HERON generates a high-resolution, obstacle-aware RM in real time for any arm supported by \texttt{cuRobo}, after training on that arm. Geometric methods cannot be used to reach this goal due to the computational cost of IK and collision checking at fine resolution. Instead, a 3D U-Net model is trained on a numerically generated dataset with a combination of binary cross-entropy (BCE) and Dice losses. This section is organized around these two aspects: the training dataset and the 3D U-Net model training.

\subsection{Training Dataset}

\texttt{cuRobo} by Sundaralingam et al.~\cite{Balakumar2023} provides GPU-accelerated batched IK solving with self-collision and world-obstacle awareness, using an ESDF representation~\cite{millane_nvblox_2024} that enables collision checking at resolutions down to 1 cm. Unlike IKFast~\cite{diankov_automated_2018}, which does not consider environment collisions, \texttt{cuRobo} integrates obstacle awareness directly into the IK pipeline. For this work, a resolution of 2 cm is used, though 1 cm or finer resolutions are achievable depending on available GPU resources.

The dataset generation follows the sphere-based sampling strategy introduced by Zacharias~\cite{Zacharias} and later adopted by Reuleaux~\cite{Makhal2018}: for each voxel of the robot's workspace, $N = 50$ poses are generated uniformly distributed on an inscribed sphere, pointing toward its center. For each pose, an IK solution is requested from \texttt{cuRobo}, and the voxel receives the reachability score $R_i \in [0, 1]$ of Eq.~\eqref{eq:reachability_index}. The choice of 50 poses per voxel follows the discretization used in Reuleaux~\cite{Makhal2018}, providing a balance between angular resolution and computational cost. To reduce the number of IK queries, we adopt the preprocessing strategy of Inv-Reach-Net~\cite{sandakalum_inv-reach_nodate}: we precompute a Master IK file containing all poses reachable by the robot in an obstacle-free environment. Poses outside this set are discarded before dataset generation, significantly reducing the number of IK queries without loss of accuracy.

\begin{figure}[t]
    \centering
    \includegraphics[trim={0 0 0 3cm}, clip, width=1.0\linewidth]{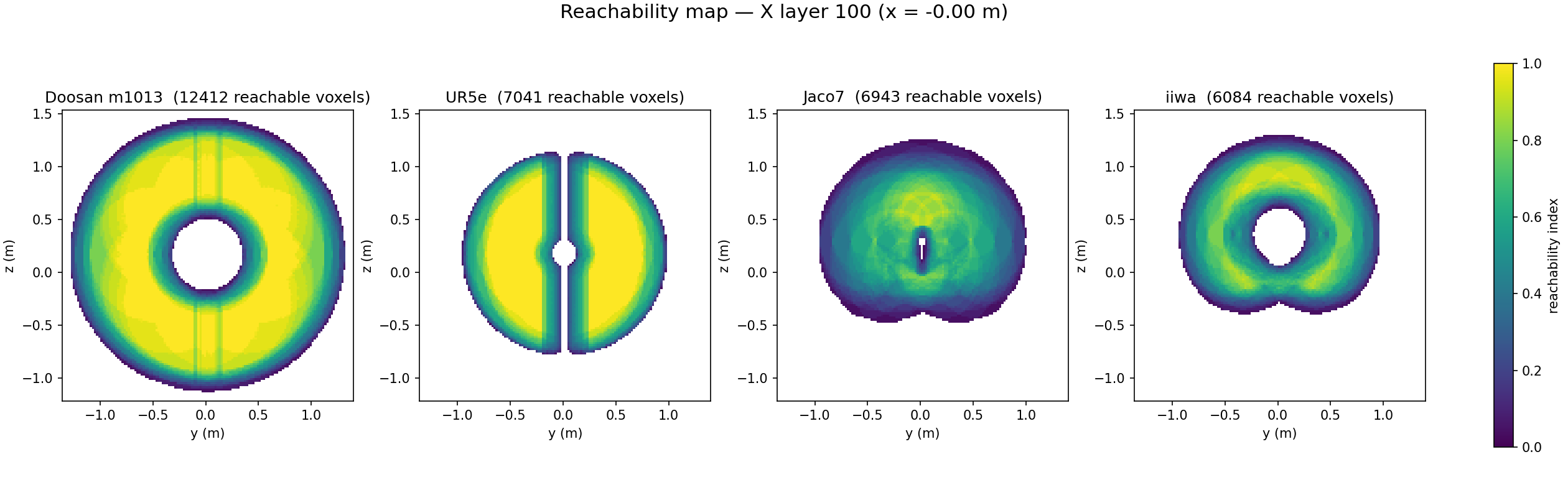}
    \caption{Ground-truth RMs of four arms computed with \texttt{cuRobo}, without obstacles, on a $152^3$ grid at 2 cm resolution (vertical slice in the plane $x=0$). From left to right: Doosan M1013, UR5e, Jaco7 and iiwa. Color gives the reachability score $R_i$, from 0 (no reachable orientation) to 1 (all 50 orientations reachable).}
    \label{fig:reachability_map}
\end{figure}

We first applied HERON to a Doosan M1013, a 6-DOF arm with a spherical wrist and a reach of 1.3 m. At 2 cm resolution, the $128^3$ grid contains 2,097,152 voxels, corresponding to 104,857,600 poses to evaluate before filtering. With the Master IK preprocessing, generating the RM of one training sample at 2 cm resolution takes approximately 90 minutes. Figure~\ref{fig:reachability_map} shows obstacle-free RMs computed in the same way for this arm and three others.

We train the network in two stages: pre-training on synthetic data, then fine-tuning on real data. The first stage uses a dataset generated at 2 cm resolution with random obstacles. The second stage fine-tunes the network on real-world data at 2 cm resolution. ROS bags were recorded with a depth camera positioned alongside the robot, capturing both the scene and the robot. The camera's depth map is passed to \texttt{cuRobo} to generate the ESDF. Because the robot itself is visible in the camera's field of view, a mesh-based segmentation step removes the robot from the collision map before dataset generation.

Datasets are stored in HDF5 format. The RM and voxel map are saved as cubic arrays of $128^3$ voxels, and metadata stores the origin, resolution, and array dimensions. At 2 cm resolution, a single sample occupies approximately 500 kB. This pipeline produces approximately 1,000 samples across the two stages, covering a range of obstacle configurations sufficient to train the network described in the following section.

\subsection{Network Architecture}

Selecting an appropriate neural network architecture for HERON requires satisfying three constraints simultaneously: real-time inference on embedded hardware, limited training data, and dense per-voxel prediction over a regular 3D grid. These constraints progressively eliminate candidate architectures.

Transformers are first excluded: despite their strong performance on volumetric tasks, they require significantly more training data and GPU memory than available on the target hardware, both for training and inference. Among 3D architectures, the input is a regular voxel grid and the output has the same spatial shape, a property that directly constrains the architecture. PointNet++~\cite{qi_pointnet_2017} is excluded as it operates on unstructured point clouds rather than regular grids. VoxelNet~\cite{noauthor_voxelnet_nodate} is designed for sparse object detection via bounding boxes and is not suited for dense per-voxel value prediction. This leaves encoder-decoder convolutional neural networks (CNNs) as the natural candidate. Among these, 3D U-Net~\cite{cicek_3d_2016} is preferred over a plain encoder-decoder thanks to its skip connections, which preserve fine spatial details during decoding, critical when predicting reachability scores at 2 cm resolution near obstacles.

\begin{figure}[t]
    \centering
    \resizebox{\linewidth}{!}{\input{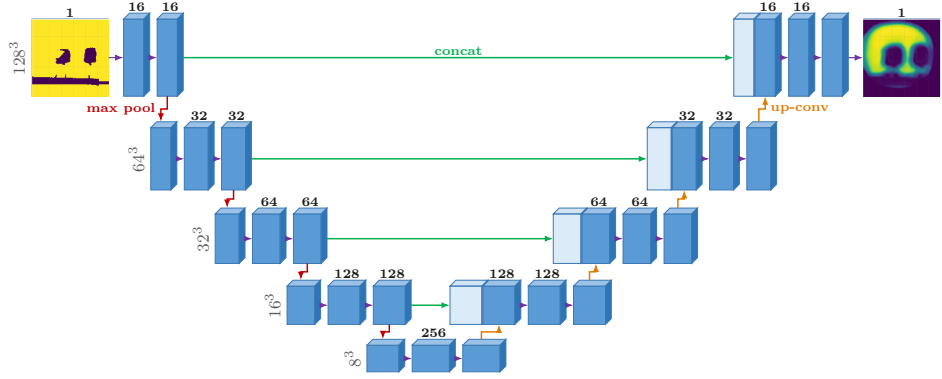}\unskip}
    \caption{HERON's architecture, a 3D U-Net. A $128^3$ obstacle map is downsampled by four max-pooling stages (16 to 256 feature maps), then upsampled by transposed convolutions to an RM of the same size. Skip connections (green) preserve fine spatial detail; each block applies convolution, normalization, and activation (purple arrows).}
    \label{fig:unet3d_diagram}
\end{figure}

The problem is formulated as regression rather than classification. A classification model would predict one of 51 discrete classes $\{0, 1, \ldots, 50\} \subset \mathbb{N}$, corresponding to the 51 possible reachability scores, requiring 51 output channels. A regression model predicts a continuous value $R_i \in [0, 1]$ with a single output channel, reducing inference time and memory footprint with no operational loss, since continuous scores are sufficient for ranking and selecting base positions. HERON therefore adopts a single-channel regression formulation, illustrated by Fig.~\ref{fig:unet3d_diagram}.

\subsection{Loss Function}
The loss function is a combination of BCE and Dice Loss, following recommendations from the segmentation literature \cite{terven_comprehensive_2025}. BCE provides a local, per-voxel gradient signal that aligns predicted probabilities with ground-truth reachability scores, and maintains strong gradients in regions where scores are near 0 or 1, where the RM is most densely populated. However, BCE treats each voxel independently, whereas an RM combines a large unreachable region, about half of the volume, with a thin transition shell at the workspace boundaries. Dice Loss addresses this by directly optimizing the spatial overlap between the predicted and ground-truth RM, penalizing inadequate coverage of the reachable workspace as a whole rather than pixel by pixel. This combination of local precision from BCE and global structure from Dice is consistent with established practice in volumetric prediction tasks \cite{terven_comprehensive_2025, cicek_3d_2016}.

\subsection{Validation Metrics}

The reachability score $R_i \in [0, 1]$ has a dual nature: it is a continuous regression output, but each value corresponds to a discrete number of reachable poses out of 50. Both regression and classification metrics are therefore used to provide a comprehensive evaluation.

For regression, the MAE and RMSE measure per-voxel prediction accuracy, with RMSE penalizing large errors more heavily. The Pearson correlation coefficient $r$ measures the linear correlation between predicted and ground-truth score distributions, providing a global measure of structural agreement between the two maps.

For classification, each continuous prediction is mapped to the nearest class among the 51 possible values $\{0, \frac{1}{50}, \ldots, \frac{50}{50}\}$, framing the evaluation as an ordinal classification problem. For such problems, the Quadratic Weighted Kappa (QWK) is preferred over global accuracy: the latter is dominated by unreachable voxels (class 0), which represent approximately 52\% of the total volume and are trivially predicted correctly. The QWK measures the level of agreement between the predicted and ground-truth labels on an ordinal scale, penalizing disagreements proportionally to their squared distance; a value of 1 indicates perfect agreement and 0 chance-level agreement, with values above 0.8 generally considered near-perfect~\cite{landis_measurement_1977}. To provide a finer diagnostic evaluation, accuracy is additionally computed over two sub-regions.

The Reachable region ($R_i \geq \frac{1}{50}$) contains all voxels where at least one end-effector orientation is accessible, corresponding to the interior of the robot workspace. This region excludes the trivially easy class-0 majority and focuses evaluation on meaningful workspace predictions.

The Ridge region contains the top 1\% of voxels ranked by local gradient magnitude $|\nabla R_i|$, computed as the 3D finite difference of the ground-truth reachability map. These voxels correspond to abrupt spatial transitions: the outer and inner boundaries of the workspace donut, kinematic singularity lines, and obstacle edges. Although this region represents only $\sim$1\% of the total volume, it is the most challenging to predict accurately due to the well-known spectral bias of convolutional networks toward low-frequency components~\cite{rahaman_spectral_2019}, which tends to smooth these sharp transitions.

These metrics correspond to the third validation level identified in the related work: quantitative evaluation against a computational ground truth generated by \texttt{cuRobo}. Validation on a real robot in cluttered environments, which would constitute a fourth level, remains an avenue for future work due to the difficulty of obtaining statistically representative ground truth under real sensing conditions.

\section{Experimental Setup}
\label{sec:ExperimentalSetup}

The target platform consists of a 6-DOF Doosan M1013 arm mounted on a Clearpath Ridgeback omnidirectional mobile base, connected via a rigid structure. An Azure Kinect 3D depth camera is positioned on the upper side of the arm, as shown in Fig.~\ref{fig:setup_robot}. All onboard computation is handled by an NVIDIA Jetson AGX Orin with 64 GB of shared memory. The software stack is built on ROS2 Humble and deployed via Docker.

This section is divided into two parts: dataset generation and 3D U-Net training, followed by real-time implementation and deployment on the MM.

\subsection{Dataset Generation and Training}

We generated the datasets on an NVIDIA H100 GPU with 60 GB of VRAM, using \texttt{cuRobo}'s GPU-accelerated batched IK pipeline. Generating one sample at 2 cm resolution took approximately 90 minutes.

The first dataset was generated for the arm only, without the mobile base or camera in the scene, allowing the model to learn the robot's kinematic geometry independently. The second dataset incorporated real depth map images with both the mobile base and the camera as obstacles, fine-tuning the model to real sensor conditions. This two-stage approach enables the method to generalize to different MM configurations sharing the same arm. Adapting HERON to a different arm morphology would require retraining from scratch, as the kinematic geometry encoded in the first stage is robot-specific.

Training was conducted with inference-time constraints in mind. While Group Normalization~\cite{wu_group_2018} is generally recommended for small batch sizes due to its independence from batch statistics, it cannot be fused with convolutional operations into a single CUDA kernel at inference time. Batch Normalization (BN), by contrast, allows Conv+BN+activation to be folded into a single kernel by TensorRT (TRT), as the running statistics become compile-time constants after training. The layer order Conv--BN--ReLU is therefore adopted throughout the network, as depicted in Fig.~\ref{fig:unet3d_diagram}.

Training 3D volumetric data with BN introduces a memory constraint: BN requires sufficiently large batch sizes to estimate stable statistics, but loading multiple $128^3$ voxel grids simultaneously is memory-intensive. On an NVIDIA A100 (40 GB), a batch size of 4 with samples loaded one at a time was the largest configuration that fit within the memory budget while maintaining stable BN statistics. Among patch sizes satisfying the architectural constraint of divisibility by $2^d$, $128^3$ is the largest that fits this configuration. At 2 cm resolution, it covers a cube of $2.56 \times 2.56 \times 2.56~\mathrm{m}^3$ centred on the arm base, i.e., 1.28 m on each side of the base, only 2 cm short of the arm's 1.30 m maximum reach.

\subsection{Real-Time Inference}
\label{sec:Inference}

Achieving real-time performance requires optimizing the full pipeline: perception preprocessing, ROS communication, and model inference.

The Azure Kinect depth camera publishes at 30 fps. Robot self-filtering is performed by \texttt{cuRobo} using the joint state and the meshes defined in the robot URDF, removing the robot body from the depth map before ESDF generation. The resulting ESDF defines the obstacle map used as model input.

Serializing a full $128^3$ voxel grid as a ROS message would exceed practical network bandwidth and introduce significant serialization latency. Instead, a lightweight message is published at 10 Hz containing only the indices of occupied voxels and the map dimensions. A preprocessing step reconstructs the full voxel grid from this sparse representation, assigning 0 to occupied voxels and 1 to free voxels. This operation adds approximately 5 ms to the pipeline.

The model is compiled and optimized using TRT. The TRT engine relies on four elements. First, the patch size of $128^3$ satisfies TRT's kernel fusion requirements: spatial dimensions must be divisible by $2^d$ (where $d=4$ is the number of pooling levels) to avoid data-dependent interpolation nodes that prevent Conv--BN--ReLU fusion into a single CUDA kernel. Additionally, 128 is a multiple of 16, which aligns with NVIDIA Tensor Core tile sizes. Second, FP16 precision reduces memory bandwidth and accelerates matrix operations with minimal accuracy loss. Third, INT8 quantization further reduces inference time but requires a calibration dataset; validation samples were used for this purpose. Finally, CUDA Graphs capture the full execution graph as a single graph launch, eliminating the per-operation CPU-GPU synchronization overhead that would otherwise dominate short inference passes.

\section{Results}\label{sec:Results}  
This section presents the results and compares the different inference optimization levels (FP32, FP16, FP16 TRT, and INT8 TRT). The results are divided into two parts: prediction accuracy (regression and classification) and real-time performance. In the first part, the model is evaluated in float32 precision without inference-time optimization, on an evaluation dataset of 100 samples never seen during training. In the second part, the inference optimizations are evaluated in terms of prediction time and their impact on accuracy.

\subsection{Prediction Accuracy}
\begin{figure}[t]
    \centering
    \includegraphics[trim={0 0 0 1.5cm}, clip, width=0.6\linewidth]{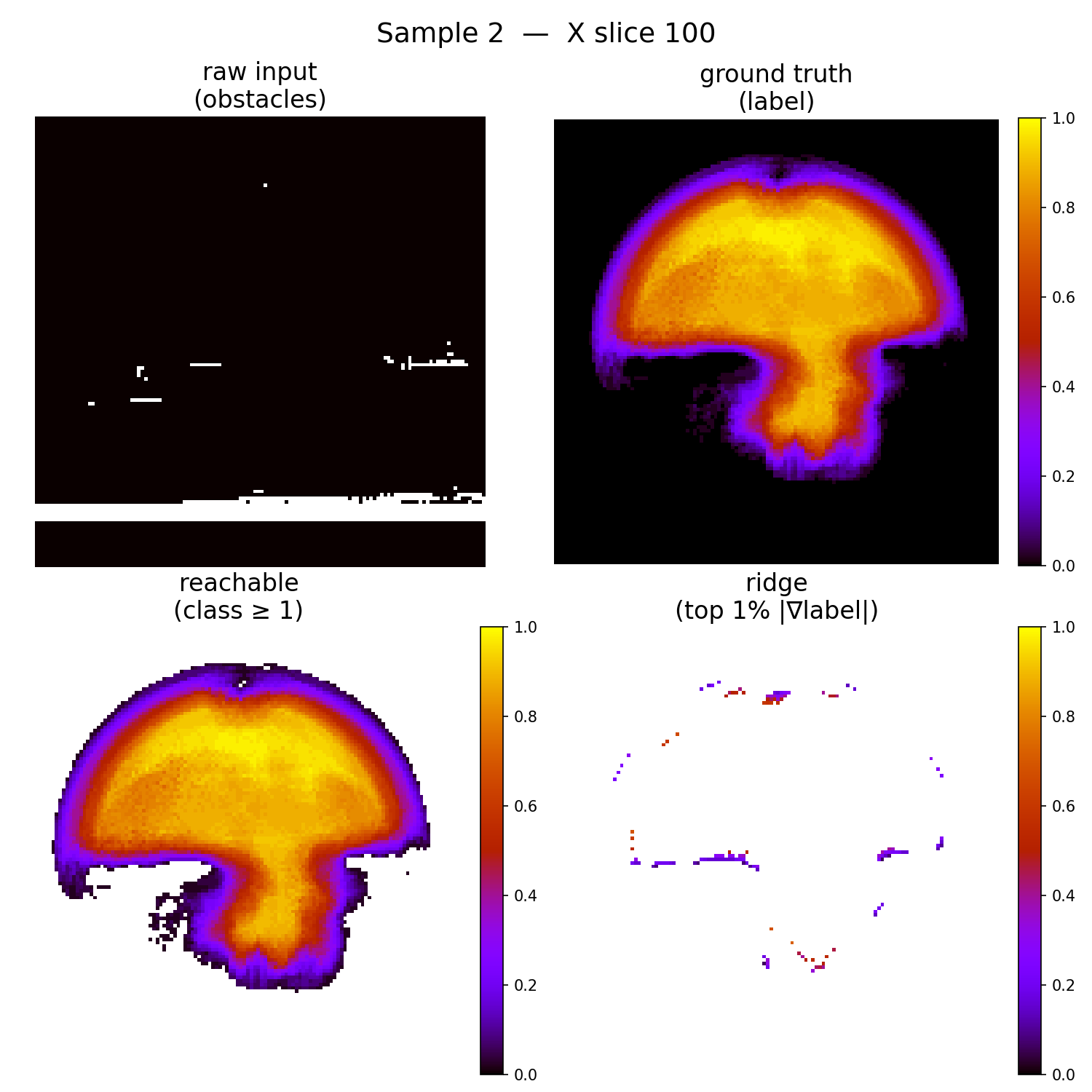}
    \caption{Visualization of the evaluation regions on an example generated with \texttt{cuRobo}: input obstacle map (top-left), ground-truth RM (top-right), reachable region (bottom-left, $R_i \geq \frac{1}{50}$), and ridge region (bottom-right, top 1\% of voxels by gradient magnitude).}
    \label{fig:prediction_regions}
\end{figure}

Figure~\ref{fig:prediction_regions} illustrates the evaluation regions on one example. Table~\ref{tab:accuracy_metrics} reports the metrics computed over the whole evaluation dataset. The Pearson correlation coefficient remains high across all regions, from 0.99 globally down to 0.98 in the ridge region. The MAE ranges from 0.6\% of the reachability score globally to 2.0\% in the ridge region, and the RMSE from 1.5\% to 3.9\%. This pattern suggests accurate global prediction while revealing a clear degradation in the ridge region, the high-gradient transitions near workspace boundaries. This degradation is consistent with the spectral bias of convolutional networks~\cite{rahaman_spectral_2019}, which preferentially smooth high-frequency spatial components such as the sharp transitions of the ridge region.

\begin{table}[t]
\caption{Result of regression metrics (MAE, RMSE, Pearson $r$) and ordinal metric (QWK) obtained with the FP32 3D U-Net model, across evaluated workspace regions.}
\centering
\label{tab:accuracy_metrics}
\begin{tabular}{@{}llccc@{}}
\toprule
Metric & & Global & Reachable & Ridge \\
\midrule
\multirow{2}{*}{MAE}         & Mean & 0.00657  & 0.0148  & 0.0207 \\
                             & Std  & 0.00108  & 0.00251 & 0.00747 \\
\midrule
\multirow{2}{*}{RMSE}        & Mean & 0.0154   & 0.0231  & 0.0393 \\
                             & Std  & 0.00403  & 0.00605 & 0.0186 \\
\midrule
\multirow{2}{*}{Pearson $r$} & Mean & 0.999    & 0.997   & 0.980 \\
                             & Std  & 0.000585 & 0.00165 & 0.0216 \\
\midrule
QWK                          & Mean & 0.999    & 0.997   & 0.980 \\
\bottomrule
\end{tabular}
\end{table}

\begin{figure}[t]
    \centering
    \includegraphics[width=0.75\linewidth]{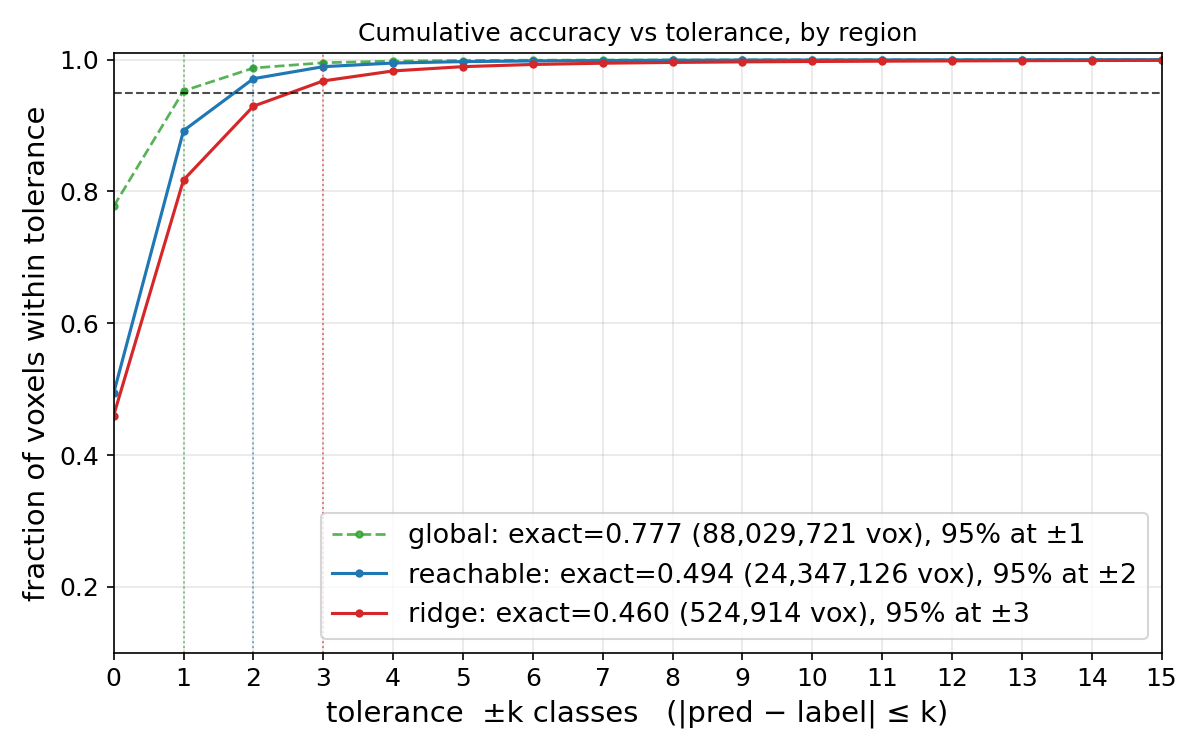}
    \caption{Fraction of voxels whose predicted class lies within $\pm k$ classes of the ground truth, for the global, reachable and ridge regions (FP32 model). The dashed line marks 95\%; dotted lines mark the tolerance at which each region reaches it.}
    \label{fig:tolerance_curve}
\end{figure}

The ordinal nature of the RM calls for classification metrics that account for class ordering. Exact-match accuracy is intentionally severe: with 51 ordinal classes, 77.7\% of voxels are predicted exactly across the full volume. However, this global figure is inflated by the trivially predicted empty class (class 0), which represents roughly half the volume. Restricted to the reachable region, exact-match accuracy drops to 49.4\%, and further to 46.0\% in the ridge region, reflecting the difficulty of predicting fine ordinal values in high-gradient zones. This metric is misleading for an ordinal target, as it penalizes near-misses as heavily as gross errors. Introducing a tolerance shows that errors are in fact small, as shown in  Fig.~\ref{fig:tolerance_curve}: within $\pm$1 class, accuracy reaches 95\%, requiring $\pm$3 classes in the ridge region. The QWK, which penalizes errors proportionally to their squared distance, is consistent with this result: it reaches 0.99 globally and 0.98 in the ridge region, above the 0.8 threshold for near-perfect agreement. The model thus captures the ordinal structure even in high-gradient regions.

\subsection{Real-Time Performance}

This section evaluates the impact of inference optimizations on accuracy and on two time measures, both obtained on the Jetson AGX Orin onboard the robot. The prediction time $t_p$ is the duration of one inference of the network alone, averaged over 100 samples. The system rate $f_s$ is the rate at which new RMs are published on the output ROS topic, measured with \texttt{ros2 topic hz}; it therefore covers the whole processing chain, from the reception of the voxel grid to the publication of the RM. As shown in Table~\ref{tab:inference_performance}, $t_p$ decreases from 872.2 ms for the unoptimized FP32 model to 115.0 ms for the INT8 TRT model. The most significant gain comes from TRT kernel fusion and CUDA Graph capture: moving from FP16 to FP16 TRT reduces the inference time by a factor of 3.9 (575.5 ms to 146.1 ms) with no measurable loss in accuracy. INT8 quantization further reduces it to 115.0 ms. Although this introduces a slight degradation (MAE 0.006 to 0.014, tolerance widening from $\pm$1 to $\pm$3 classes), we retained INT8 for the real-time deployment.

The full pipeline combines an asynchronous voxel grid published on a ROS topic at 10 Hz with a sequential preprocessing step (5 ms) executed immediately before each prediction. The processing path (preprocessing and inference) therefore constitutes the critical path, of about 120 ms with the INT8 model. The measured system rate $f_s = 8.57$ Hz, i.e., one RM every 117 ms, is consistent with this critical path. Because $t_p$ and $f_s$ come from separate measurements, $1/f_s$ can fall slightly below $t_p$, as for FP32 (847 ms versus 872 ms), a gap well within the standard deviation of $t_p$ (160 ms). Inference thus remains the bottleneck, while the voxel-grid publication imposes a higher ceiling of 10 Hz, and the camera (30 fps) a higher one still. This suggests that the current pipeline is not saturated: further inference optimization could raise the system rate up to the 10 Hz of the voxel-grid publication. For comparison, computing the ground-truth RM of one sample with \texttt{cuRobo} takes about 90 min on an H100 GPU, while HERON predicts it about 47,000 times faster, on embedded hardware.

\begin{table}[t]
\caption{Prediction time $t_p$ (network inference only), system rate $f_s$ (whole processing chain, measured on the output ROS topic) and accuracy on the Jetson AGX Orin for each inference optimization level.}
\centering
\label{tab:inference_performance}
\begin{tabular}{@{}lcccc@{}}
\toprule
 & FP32 & FP16 & FP16 TRT & INT8 TRT \\
\midrule
Prediction time $t_p$, mean (ms)     & 872.2    & 575.5    & 146.1    & 115.0 \\
Prediction time $t_p$, std (ms)      & 160.3    & 82.7     & 20.8     & 34.2 \\
System rate $f_s$ (Hz)               & 1.18     & 1.71     & 6.46     & 8.57 \\
Tolerance for 95\% accuracy (classes) & $\pm 1$ & $\pm 1$  & $\pm 1$  & $\pm 3$ \\
\midrule
MAE                                  & 0.006566 & 0.006573 & 0.006573 & 0.014251 \\
RMSE                                 & 0.015390 & 0.015389 & 0.015389 & 0.031833 \\
Pearson $r$                          & 0.999045 & 0.999044 & 0.999044 & 0.995958 \\
\bottomrule
\end{tabular}
\end{table}

\section{Discussion}
\label{sec:Discussion}

To the best of our knowledge, HERON is the first method to combine real-time generation, 2 cm resolution, and collision awareness in a 3D RM of a 6-DOF arm. ReachNet~\cite{sandakalum_reachnet_nodate} is also fast and collision-aware, but for a 3-DOF planar arm at 10 cm resolution. The generated map extends the reactive approach of Burgess-Limerick et al.~\cite{burgess-limerick_reactive_2024} to full 3D and to tasks such as cleaning. As identified in~\cite{dupoiron_adaptive_2026}, the need to reposition the base during a cleaning task differs fundamentally from the IRM-based base placement of Makhal et al.~\cite{Makhal2018}, which precomputes optimal positions from a known set of goals. In tasks such as cleaning, goals are not known in advance and may evolve during execution. Furthermore, reacting to dynamic elements, e.g., a moving human or another robot, requires re-evaluating the reachability map online, which precomputed IRM approaches cannot provide.

Two properties ease the transfer of HERON to other robots. First, the model trained in the first stage depends only on the arm: adapting HERON to a new mobile base sharing the same arm only requires fine-tuning on real data of that base. Second, training a model for a new arm requires little data: empirically, approximately 1000 samples were sufficient to reach the reported results on the Doosan M1013. The dataset generation pipeline has been applied to four arm geometries shown in Fig.~\ref{fig:reachability_map}: Doosan M1013 (6-DOF), UR5e (6-DOF), Jaco7 (7-DOF), and iiwa (7-DOF), suggesting that the generation procedure is not specific to a single morphology. 

However, the design choices introduce three limitations: spectral bias, fixed resolution, and the input representation. The spectral bias manifests as reduced accuracy in the ridge region. This limitation is inherent to neural networks, particularly CNNs, and can be mitigated but not fully eliminated. Regarding resolution, the model is trained for a fixed value, and finer resolution would increase both GPU memory and prediction time. Experiments also show that the full RM must be predicted at once: predicting only a chunk deprives the model of the global geometric context of the arm.

With HERON, using a voxel grid as input introduces two issues. First, preprocessing must convert the existing ESDF representation into a voxel grid, adding latency. Second, the current representation does not distinguish free voxels from unobserved ones, treating both identically, which can cause RM instability as the arm alternately occludes and reveals regions of the scene. 

Many of these issues could be addressed by operating directly on the ESDF. Its continuous definition would support multiple resolutions, could allow architectures beyond CNNs, enable the inclusion of additional information such as robot geometry, and inherently distinguish observed from unobserved space. 

Finally, like Sandakalum et al.~\cite{sandakalum_inv-reach_nodate} and Cavelli et al.~\cite{cavelli_modeling_2025}, this work evaluates the RM at the third validation level defined in Section~\ref{sec:RelatedWork}, i.e., quantitatively against a computational ground truth, and extends this evaluation to real depth data and to region-specific metrics. The next step would be fourth-level validation through real-robot testing. Such evaluation is difficult to set up, primarily because obtaining a reliable real-world ground truth is itself challenging, every experimental setup introduces biases that are complex to identify and correct. This is why several works, such as Makhal et al.~\cite{Makhal2018}, evaluate their methods through task success and human assessment instead. In the absence of a reliable real-world ground truth, human judgment remains a practical reference for assessing whether a pose is reachable.

\section{Conclusion}
\label{sec:Conclusion}

This article presented HERON, a method to generate high-resolution, collision-aware reachability maps in real time. The approach relies on a dataset generated using the \texttt{cuRobo} framework, and a 3D U-Net model trained to predict the RM directly from a voxelized obstacle representation. The method was validated on a 6-DOF robotic arm (Doosan M1013). Regarding the prediction rate, the deployed INT8 model runs at 8.57 Hz on a Jetson AGX Orin, close to the 10 Hz target set by the obstacle-map updates; inference remains the bottleneck. Regarding the agreement with the \texttt{cuRobo} ground truth, 95\% of the voxels lie within $\pm$1 class with the FP32 model and within $\pm$3 classes with the INT8 model, the main degradation occurring in the ridge region. HERON was validated at the third level identified in the literature, i.e., quantitatively against a computational ground truth. As for IK solvers, exhaustively testing an RM on a real robot is hardly feasible, which makes this level the most thorough validation practicable today, beyond the qualitative or comparative validation typically reported in prior work. The contributions of the article are:
\begin{itemize}
    \item a pipeline that computes collision-aware RMs at 2 cm resolution with \texttt{cuRobo}, from synthetic scenes and real depth images;
    \item a 3D U-Net that predicts the RM from a $128^3$ occupancy grid, designed for embedded deployment with TRT;
    \item an evaluation onboard an MM that reports the accuracy in three regions of the workspace and the latency of each inference optimization level.
\end{itemize}

Future work will follow two main directions. The first is to operate directly on the ESDF collision representation rather than a voxel grid, removing the preprocessing step and enabling the model to distinguish between occupied, free, and unobserved space. The second is to use the generated RM to reposition the mobile base online during task execution, guided by the predicted quality of the workspace, and to extend validation toward real-robot testing across additional arm morphologies.

\backmatter

\bmhead{Acknowledgements}
This research was enabled in part by support provided by Calcul Qu\'ebec (calculquebec.ca) and the Digital Research Alliance of Canada (alliancecan.ca).

\section*{Declarations}

\bmhead{Funding}
This work was supported by the Natural Sciences and Engineering Research Council of Canada (NSERC) through the Collaborative Research and Training Experience (CREATE) program, specifically the Collaborative Robotics for Manufacturing (CoRoM) program, and by Mitacs through the Mitacs Accelerate program.

\bmhead{Competing interests}
The authors declare no competing interests.

\bmhead{Ethics approval and consent to participate}
Not applicable.

\bmhead{Consent for publication}
Not applicable.

\bmhead{Data availability}
The datasets generated and analysed during the current study are available in the Heron repository https://github.com/Lab-CORO/Heron.

\bmhead{Code availability}
The source code used in this study is openly available at https://github.com/Lab-CORO/Heron under the GNU GENERAL PUBLIC LICENSE V3 license.

\bmhead{Author contributions}
All authors contributed to the study conception and design. Material preparation, data collection and analysis were performed by Dupoiron Guillaume. The first draft of the manuscript was written by Dupoiron Guillaume and all authors commented on previous versions of the manuscript. All authors read and approved the final manuscript.

\bibliography{Bibliography/heron}

\end{document}